\pdfoutput=1
\documentclass[runningheads]{llncs}
\usepackage{graphicx}
\usepackage{tikz}
\usepackage{comment}
\usepackage{amsmath,amssymb} % define this before the line numbering.
\usepackage{color}

\usepackage[accsupp]{axessibility}  % Improves PDF readability for those with disabilities.

\begin{document}
%\renewcommand\makeLineNumber {\hss\thelinenumber\ \hspace{6mm} \rlap{\hskip\textwidth\ \hspace{6.5mm}\thelinenumber}}
% \linenumbers
\pagestyle{headings}
\mainmatter
\def\ECCVSubNumber{14}  % Insert your submission number here

\title{Affective Behavior Analysis using Action Unit Relation Graph and Multi-task Cross Attention} % Replace with your title

\begin{comment}
\titlerunning{ECCV-22 submission ID \ECCVSubNumber} 
\authorrunning{ECCV-22 submission ID \ECCVSubNumber} 
\author{Anonymous ECCV submission}
\institute{Paper ID \ECCVSubNumber}
%\institute{Paper ID \ECCVSubNumber}

\end{comment}
%******************

%\begin{comment}
\titlerunning{AU Relation Graph and Multi-task Cross Attention in Affective Analysis}
\author{Dang-Khanh Nguyen \inst{1}\and
Sudarshan Pant \inst{1}\and
Ngoc-Huynh Ho \inst{1}\and
Guee-Sang Lee \inst{1}\and
Soo-Huyng Kim \inst{1}\and
Hyung-Jeong Yang \inst{1,2}}
\authorrunning{Dang-Khanh Nguyen et al.}
\institute{Department of Artificial Intelligence Convergence, Chonnam National University
Gwangju 61186, South Korea \and
Corresponding author: \email{hjyang@jnu.ac.kr}}
%\end{comment}
%******************
\maketitle

\begin{abstract}
Facial behavior analysis is a broad topic with various categories such as facial emotion recognition, age, and gender recognition. Many studies focus on individual tasks while the multi-task learning approach is still an open research issue and requires more research. In this paper, we present our solution and experiment result for the Multi-Task Learning challenge of the Affective Behavior Analysis in-the-wild competition. The challenge is a combination of three tasks: action unit detection, facial expression recognition, and valance-arousal estimation. To address this challenge, we introduce a cross-attentive module to improve multi-task learning performance. Additionally, a facial graph is applied to capture the association among action units. As a result, we achieve the evaluation measure of $128.8$ on the validation data provided by the organizers, which outperforms the baseline result of $30$.
\end{abstract}

\section{Introduction}
In affective computing, emotion recognition is a fundamental research topic and our face is an obvious indicator to analyze the human affect. With the development of computer vision and deep learning, there are numerous studies and modern applications related to facial behavior analysis \cite{kollias2019deep}, \cite{kollias2017recognition}, \cite{kollias2018photorealistic}. The ABAW 4th Workshop organized a competition with two challenges which are the multi-task learning (MTL) challenge involving the development of a multi-task model using facial images \cite{kollias2021affect} and the learning from synthetic data (LSD) challenge involving the use of synthetic data \cite{kollias2020deep},  \cite{kollias2020va}. We only participated in the MTL challenge and the LSD challenge is beyond the scope of this work.

The MTL challenge aims to design a model performing the following three tasks with facial image data as input:
\begin{enumerate}
    \item  Action unit detection (AU detection): a task involving a multi-label classification with 12 classes of action units that represent various movements on the subject's face.
    \item  Facial expression recognition (FER): a multi-class classification task, which involves identifying the emotion of the subjects among 8 categories: happiness, sadness, anger, fear, surprise, disgust, neutral and other state.
    \item Valance Arousal estimation (VA estimation): a regression task, which involves estimating the valence and arousal. Arousal labels are numeric representations of the excitement levels of the subject, while valence labels represent the degree of positive or negative feelings. The output is two continuous values in the range $[-1,1]$.
\end{enumerate}

This paper proposes utilizing an attention mechanism for MTL problem. By attending to the output of one specific task, the model can learn to boost the result of other related tasks. In addition, we leverage the graph-based neural network to learn the relation among action units (AUs) in the AU detection task.

\section{Related Work}
Based on the VGG-Face, Kollias \cite{kollias2021affect} devised a multi-task CNN network to jointly learn the VA estimation, AU detection, and FER. The MT-VGG model was created by adapting the original VGG-Face for multi-tasking purposes with three different types of outputs. The author also proposed the recurrent version to adapt to temporal affect variations and the multi-modal version to exploit the audio information.

Savchenko \cite{savchenko2021facial} introduced a multi-head model with a CNN backbone to resolve the facial expression and attributes recognition. The model was sequentially trained with typical face corpora \cite{sharma2019automatic}, \cite{cao2018vggface2}, \cite{mollahosseini2017affectnet} for various facial analysis tasks. In ABAW 3rd competition, Savchenko \cite{savchenko2022frame} also developed a lightweight model using EfficientNet \cite{tan2019efficientnet} to effectively learn the facial features and achieved the top 3 best performances in the MTL challenge.

Kollias \cite{kollias2021distribution} showed an association between emotions and AUs distribution. Each emotion has its prototypical AUs, which are always active along with it; and observational AUs, which are frequently present with the emotion at a certain rate. From this assumption, the authors proposed co-annotation and distribution matching to couple the emotions and AUs. Luo \cite{luo2022learning} introduced a graph-based method with multi-dimensional edge features to learn the association among AUs. The AU detection block in our model is inspired by the node feature learning in \cite{luo2022learning}. 

\section{Proposed Method}
Our method is based on two observations: (1) there are informative connections among AU activations \cite{luo2022learning} and (2) the presence of the AUs is related to the facial expression \cite{kollias2021distribution}. Following these statements, we proposed a model with a graph convolution network (GCN) to exploit the AUs’ connections and a cross attention module to learn the influence of AUs' presence on facial emotion expression.

\begin{figure}
\centering
\includegraphics[height=4.5cm]{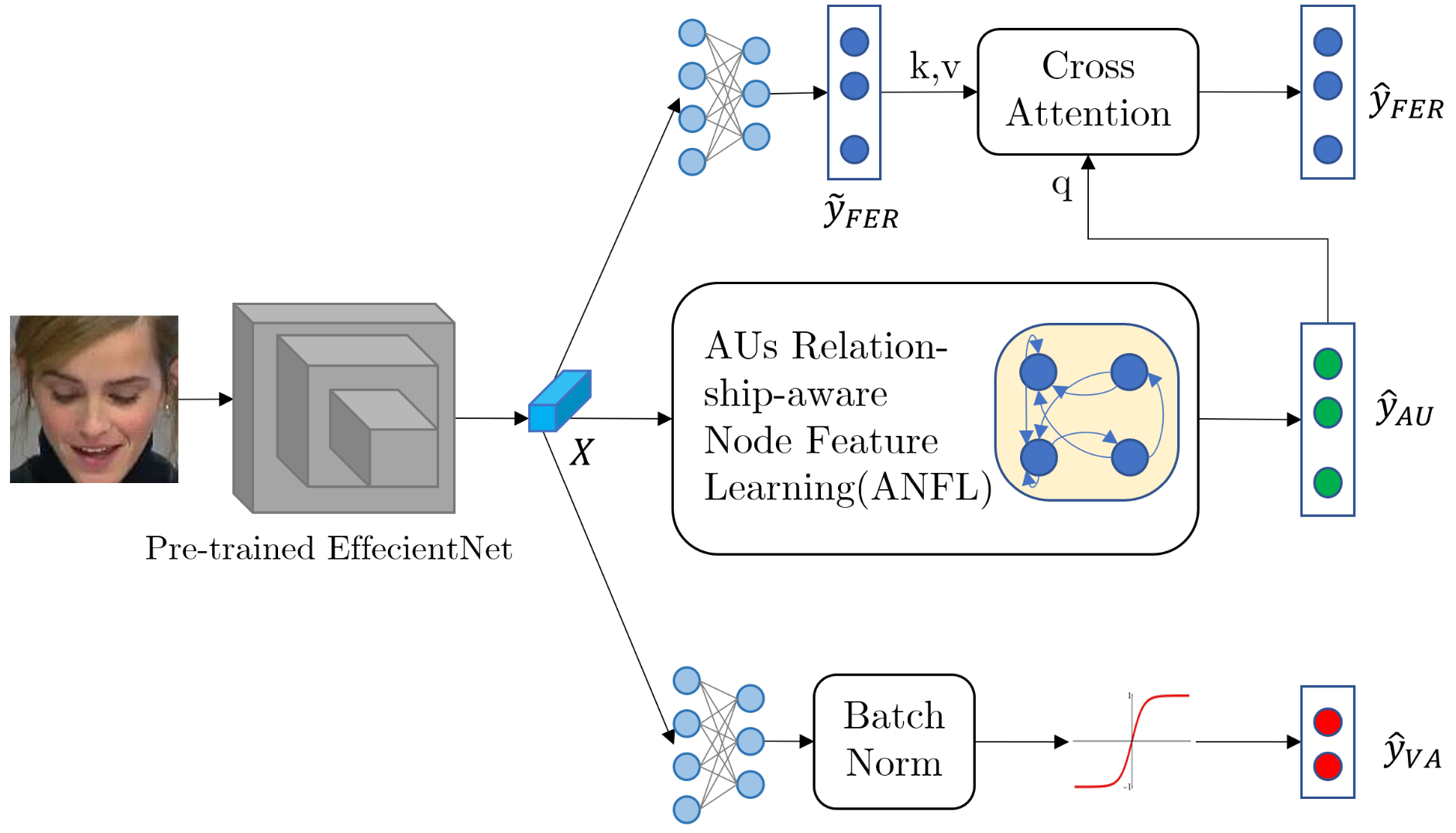}
\caption{Block diagram of our multi-task cross attention model}
\label{fig:arc}
\end{figure}

The architecture of the proposed model is illustrated in Fig~\ref{fig:arc}. We used a pre-trained EfficientNet \cite{savchenko2021facial} to obtain the facial feature vector from the input image. The image feature is then fed to three blocks corresponding to three tasks. Regarding the AU detection task, we utilized the AU relation graph to create the AU-specific features. For expression recognition and valance-arousal estimation, we used two fully connected layers to generate the predictions. In addition, we devised an attention-based module to learn the effect of facial AUs on the prediction of the emotion recognition task. 

\subsection{AU relation graph}
To learn the representation of AUs, we adopted the Action unit Node feature learning (ANFL) module proposed by Luo \cite{luo2022learning}. The ANFL module consists of $N$ fully connected layers corresponding to $N$ AUs. These layers generate the AU-specific feature vectors $\lbrace v_i \rbrace^{N}_{i=1}=V$ using the extracted image feature $X$. Mathematically, the AU-specific feature vectors are given by:
\begin{align}
  v_i = \sigma\left(W_{i}X + b_i\right)
\end{align}

Afterward, the Facial Graph Generator (FGG) constructs a fully connected graph with N nodes corresponding to $N$ AU-specific feature vectors. The edge weight of two nodes is the inner dot product of the two corresponding vectors. The graph is simplified by removing edges with low weights. We chose k-nearest neighbors algorithm to keep valuable connections between nodes. We used the simplified topology to create the adjacency matrix of a GCN. The GCN is used to enrich the connection information among AU-specific feature vectors. Generally, the AU-specific feature vectors learned by the GCN network are denoted by:
\begin{align}
  V^{FGG} = f_{FGG}\left(V\right)
\end{align}

\begin{figure}
\centering
\includegraphics[height=5.5cm]{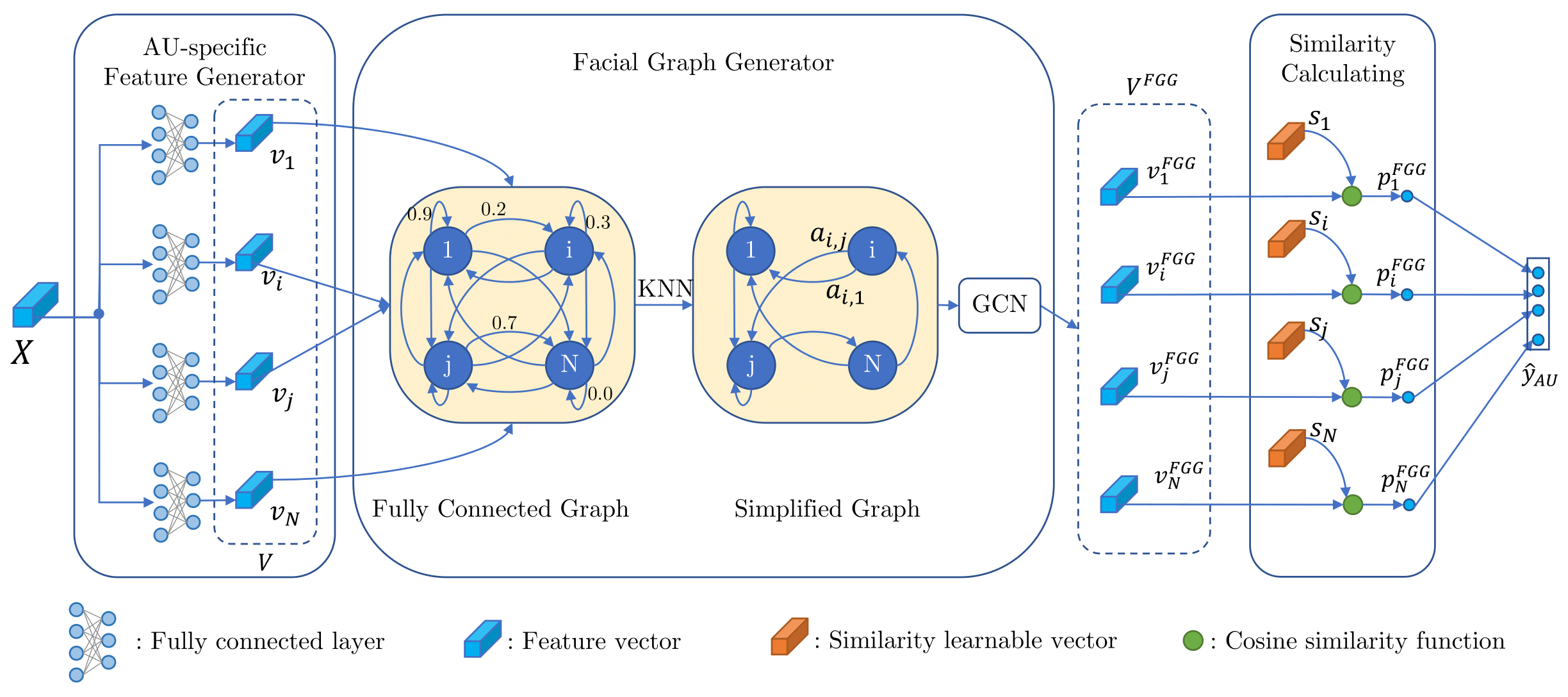}
\caption{AUs Relationship-aware Node Feature Learning module}
\label{fig:graph}
\end{figure}

Finally, we calculate the similarity between the $v^{FGG}_i$ and $s_i$ to get the probability of each AU activation using the node features from the GCN. The similarity function is defined by:
\begin{align}
  \hat{y}_{AU,i} = \frac{ReLU\left(v^{FGG}_i\right)^T ReLU\left(s_i\right)}{{\lVert ReLU\left(v^{FGG}_i\right)\rVert}_2 {\lVert ReLU\left(s_i\right)\rVert}_2}
\end{align}
where $s_i$ is trainable vector having same dimension as $v^{FGG}_i$.
The operation of ANFL is illustrated in Fig~\ref{fig:graph}. More detail about ANFL is in \cite{luo2022learning}.

\subsection{FER and VA estimation heads}
To estimate FER and VA, we simply put the image feature X into two fully connected layers in parallel. We used Batch Normalization and Tanh activation function to produce the valance-arousal value $\hat{y}_{VA}$. Meanwhile, FER head generates unweighted logit prediction $\tilde{y}_{EX}$ without any activation function. With trainable parameters $W_{VA}, b_{VA}, W_{EX}, b_{EX}$, the formulas of $\hat{y}_{VA}$ and $\tilde{y}_{EX}$ are given below:
\begin{align}
  \hat{y}_{VA} = \tanh\left(W_{VA}X + b_{VA}\right) \; \\
  \tilde{y}_{EX} = W_{EX}X + b_{EX}
\end{align}

Inspired by the additive attention of Bahdanau \cite{bahdanau2014neural}, we devised a multi-task cross attention module to discover the relationship between AU prediction and facial expression. Given an AU prediction $\hat{y}_{AU}$ (query) and FER unweighted prediction $\tilde{y}_{EX}$ (key), the attention module generates the attention scores as the formula below:
\begin{align}
  h = W_v * \tanh\left(W_{q}\hat{y}_{AU}+W_{k}\tilde{y}_{EX}\right)
\end{align}
where $W_{q}, W_{k}, W_v$ are trainable parameters.

The attention weights $a$ are computed as the softmax output of the attention scores $h$. This process is formulated as:
\begin{align}
  a = Softmax(h)
\end{align}

Finally, the weighted FER prediction is the element-wise product of attention weight $a$ and the FER unweighted prediction $\tilde{y}_{EX}$ (value):
\begin{align}
  \hat{y}_{EX} = a * \tilde{y}_{EX}
\end{align}

\subsection{Loss function} \label{lossfunction}
\begin{figure}
\centering
\includegraphics[height=4cm]{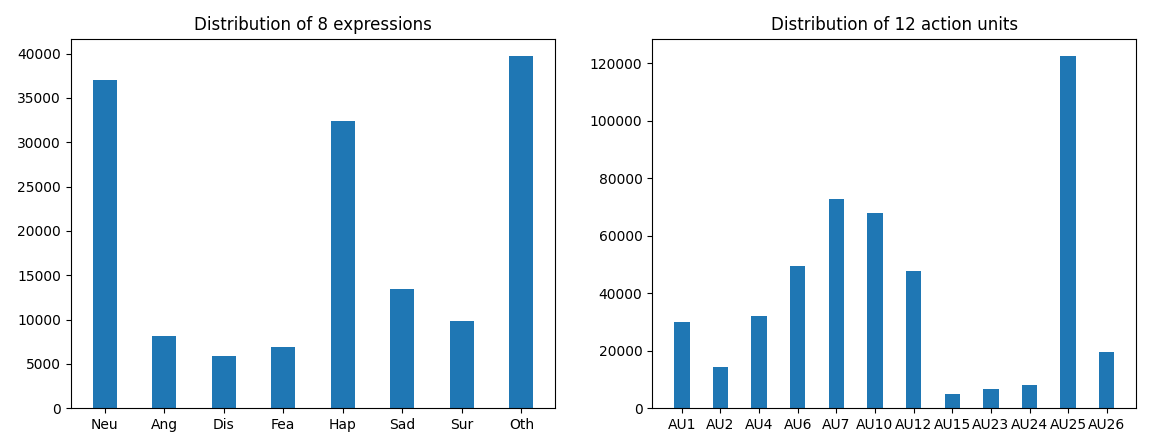}
\caption{Distribution of facial attributes in s-Aff-Wild2 database}
\label{fig:distribute}
\end{figure}

We used the weighted asymmetric loss proposed by Luo \cite{luo2022learning} for AU detection task training process:

\begin{align}
  \mathcal{L}_{AU}=-\frac{1}{N}\sum_{i=1}^{N}w_{i}\left[y_{AU,i}\log\left(\hat{y}_{AU,i}\right)+\left(1-y_{AU,i}\right)\hat{y}_{AU,i}\log\left(1-\hat{y}_{AU,i}\right)\right]
\end{align}
Each binary entropy loss of an AU is multiplied with a coefficient $w_i$. These factors are used to balance the contribution of each AU in the loss function because their appearance frequencies of AUs are different from each other as in Fig~\ref{fig:distribute} . The factor $w_i$ is computed from occurrence rate $r_i$ of AU $i^{th}$ in the dataset:
\begin{align}
  w_i=N\frac{\frac{1}{r_i}}{\sum_{j=1}^N \frac{1}{r_j}}
\end{align}

As the ratio of 8 expressions are imbalance, the loss we use for FER task is weighted cross entropy function, which is given by:
\begin{align}
  \mathcal{L}_{EX}=-\sum_{i}^{C}P_{i}y_{EX,i}\log\left({\rho}_i\left(\hat{y}_{EX}\right)\right)
\end{align}
where ${\rho}_i\left(\hat{y}_{EX}\right)$ represents the softmax function, $P_{i}$ is the refactor weight calculated from training set and $C$ is the number of facial expressions.

For VA estimation task, the loss function is obtained as the sum of individual CCC loss of valence and arousal. The formula is given by:
\begin{align}
  \mathcal{L}_{VA}=1 - CCC^{V} + 1 - CCC^{A}
\end{align}
$CCC^{i}$ is the concordance correlation coefficient (CCC) and $i$ could be $V$ (valence) or $A$ (arousal). CCC is a metric measures the relation of two distributions, denoted by:
\begin{align}
  CCC = \frac{2s_{xy}}{s^{2}_x + s^{2}_y  + \left(\overline{x}-\overline{y}\right)^2}
\end{align}
where $\overline{x}$ and $\overline{y}$ are the mean values of ground truth and predicted values, respectively, $s_x$ and $s_x$ are corresponding variances and $s_{xy}$ is the covariance value.

\section{Experiment}

\subsection{Dataset}
The experiment is conducted on the s-Aff-Wild2 dataset \cite{kollias2022abaw}, which is a static version of Aff-Wild \cite{zafeiriou2017aff} and Aff-Wild2 \cite{kollias2019expression} databases. s-Aff-Wild2 is a collection of 221,928 images annotated with 12 action units, 6 basic expressions, and 2 continuous emotion labels in valence and arousal dimensions. In the s-Aff-Wild2 dataset, some samples may lack annotations for one of the three mentioned labels. Such missing labels may cause an imbalance in the number of valid annotations among the batches split by the data sampler. For simplicity, instead of implementing a dedicated data sampler, we decided to train three tasks separately.

The organizers provided the facial cropped images extracted from the videos and corresponding cropped-aligned images. We only used the cropped-aligned version to exploit the facial features. We resized all images from $112 \times $112 to $224 \times $224 pixel, normalized, applied random cropping and random horizontal flipping before the feature extraction step.

\subsection{Evaluation metrics}
Following \cite{kollias2022abawlsd}, the evaluation metric of the MTL challenge $\mathcal{P}_{MTL}$ is the summation of three uni-task performance measure:
\begin{align}
  \mathcal{P}_{MTL} = \mathcal{P}_{AU} + \mathcal{P}_{EX} + \mathcal{P}_{VA}
\end{align}
where $\mathcal{P}_{AU}$ is the average F1 score of the 12 AUs in AU detection task, $\mathcal{P}_{EX}$ is the average F1 score of 8 expression categories in FER task and $\mathcal{P}_{VA}$ is the average of CCC indexes of valance and arousal in VA estimation task. The performance metrics of individual tasks can be formulated as:
\begin{align}
  \mathcal{P}_{AU} =\frac{\sum_{i}^{N} {F}^{AU,i}_1}{12}\\
  \mathcal{P}_{EX} =\frac{\sum_{i}^{C} {F}^{exp,i}_1}{8} \\
  \mathcal{P}_{VA} = \frac{CCC_V + CCC_A}{2}
\end{align}
where $CCC_V, CCC_A$ are the CCC of valence and arousal, respectively; ${F}^{exp,i}_1$ is the F1 score of the emotion label and ${F}^{AU,i}_1$ is the F1 score of the action unit.
\subsection{Experiment setting}
We implemented our solution using Pytorch framework and conducted the experiments on NVIDIA RTX 2080 Ti GPU. Stochastic Gradient Descent was applied following with Sharpness-aware Minimization Optimizer \cite{foret2021sharpnessaware} to minimize the loss function. The model was trained with an initial learning rate of $10^{-3}$ and the cosine decay learning rate scheduler was also used. Because the CCC lost function requires a sufficient sequence of predictions \cite{kollias2021distribution} so we chose the batch size of 256 for all tasks' training process. The weight decay was applied to prevent overfitting.

The pre-trained EffecientNet in \cite{savchenko2021facial} can capture facial features efficiently. Therefore, we decide not to train it in our experiment. Firstly, we train the ANFL module with AU detection tasks so that the model can learn the knowledge of AUs activation. Subsequently, we continue training the model with the FER task to take advantage of AU detection results to support the prediction of emotion. The VA estimation head can be trained independently.

\subsection{Result}

We attain the evaluation measure of $128.8\%$ compared to $30.0\%$ of the baseline model \cite{kollias2022abawlsd}. A model which uses pre-trained EfficientNet and three fully connected layers accomplishes the performance score of $121.5\%$.  By utilizing the multi-task cross attention and GCN, we improve the prediction of three tasks as shown in Table~\ref{table:result}. Our best model obtains the score of $111.35\%$ on the test set of the s-Aff-Wild2 database.  All metrics in this section are in percentage (\%). 
\setlength{\tabcolsep}{4pt}
\begin{table}
\begin{center}
\caption{The uni-task and multi-task performance of our model and other options. The evaluation metrics are performed on the validation set of the s-Aff-Wild2 dataset}
\label{table:result}
\begin{tabular}{lllll}
\hline\noalign{\smallskip}
Model & $\mathcal{P}_{AU}$ & $\mathcal{P}_{EX}$ & $\mathcal{P}_{VA}$ & $\mathcal{P}_{MTL}$ \\
\noalign{\smallskip}
\hline
\noalign{\smallskip}
Baseline  & - & - & - & 30.0\\
FC-layer heads & 41.4 & 32.2 & 47.9 & 121.5\\
FC-layer heads (BatchNorm VA head) & 45.6 & 32.2 & 47.9 & 125.7\\
\noalign{\smallskip}
\hline
\noalign{\smallskip}
Proposed method (w/o cross attention) & 45.6 & 32.2 & 49.9 & 127.7\\
Proposed method (cross attention) & {\bf 45.6} & {\bf 33.3} & {\bf 49.9} & {\bf 128.8}\\
\hline
\end{tabular}
\end{center}
\end{table}

By discovering the association of AUs, the model increases the average F1 score to $49.9$, better than the measure of $47.9$ when using a fully connected layer followed by a sigmoid activation. The SAM optimizer improves model generalization and yields better performance. The detailed results are listed in Table~\ref{table:auresult}.
\setlength{\tabcolsep}{2.5pt}
\begin{table}
\begin{center}
\caption{The comparison F1 score of each AU prediction on validation set between FC-Sig (a fully connected layer followed by a sigmoid function), ANFL, and sANFL (ANFL with SAM optimizer)}
\label{table:auresult}
\begin{tabular}{llllllllllllll}
\hline\noalign{\smallskip}
 & \multicolumn{12}{c}{AU\# }\\
Model & 1 & 2 & 4 & 6 & 7 & 10 & 12 & 15 & 23 & 24 & 25 & 26 & $\mathcal{P}_{AU}$\\
\noalign{\smallskip}
\hline
\noalign{\smallskip}
FC-Sig & 53.1 & 42.2 & 56.4 & 56.3 & 73.5 & {\bf 72.0} & 63.6 & 9.9 & 14.8 & 9.2 & 87.6 & 35.5 & 47.9 \\
ANFL & 51.2  & 36.0  & 55.8  & {\bf 56.9}  & 72.9  & 70.6  & 63.7  & {\bf 18.8}  & {\bf 15.9}  & 10.3  & 87.6  & 32.6 & 47.7 \\
sANFL & {\bf 56.6} & {\bf 42.9} & {\bf 61.1} & 55.7 & {\bf 73.7} & 71.1 & {\bf 66.5} & 18.5 & 15.7 & {\bf 13.0} & {\bf 87.8} & {\bf 36.4} & {\bf 49.9} \\
\hline
\end{tabular}
\end{center}
\end{table}
\setlength{\tabcolsep}{1.4pt}

In the VA estimation task, our assumption is that the batch normalization can boost the accuracy of the prediction. The batch normalization layer can change the mean and variance of a batch, respectively, to the new values $\beta$ and $\gamma$, which are learnable parameters. By using CCC for the loss function as in section \ref{lossfunction}, the network can learn to adapt the parameters $\beta, \gamma$ to the mean and variance of ground truth distribution. Thus, the performance of the VA estimation task can be increased. We conducted the experiments with and without the batch normalization layer to test its operation. As the result, the batch normalization layer can improve $\mathcal{P}_{VA}$ by more than 4\%. The CCC indexes of valence, arousal, and their average value are described in Table~\ref{table:varesult}.
\setlength{\tabcolsep}{4pt}
\begin{table}
\begin{center}
\caption{The VA evaluation metrics on validation set of the model with and without Batch Normalization}
\label{table:varesult}
\begin{tabular}{llll}
\hline\noalign{\smallskip}
Model & ${CCC}^{V}$ & ${CCC}^{A}$ & $\mathcal{P}_{VA}$ \\
\noalign{\smallskip}
\hline
\noalign{\smallskip}
FC+Tanh Activation & 43.9 & 38.8 & 41.4\\
FC+BatchNorm+Tanh Activation & {\bf 47.5} & {\bf 43.6} & {\bf 45.6}\\
\hline
\end{tabular}
\end{center}
\end{table}
\setlength{\tabcolsep}{1.4pt}

In the FER network, there is a noticeable improvement when we use cross attention. We obtain the average F1 score of 33.3 which is considerably higher than the case of not using multi-task cross attention. To evaluate the contribution of the attention module, we compared the F1 scores for individual emotion labels by excluding the attention mechanism. As shown in Table~\ref{table:exresult}, the use of attention increased the performance for individual labels except for \textit{happy} and \textit{surprise}. 

\begin{table}
\begin{center}
\caption{The F1 scores of emotion labels on validation set of the model with and without cross-attention}
\label{table:exresult}
\begin{tabular}{llllllllll}
\hline\noalign{\smallskip}
Model & Neu & Ang & Dis & Fea & Hap & Sad & Sur & Oth & $\mathcal{P}_{EX}$\\
\noalign{\smallskip}
\hline
\noalign{\smallskip}
Without attention & 35.7 & 24.0 & 45.4 & 23.3 & {\bf 43.9} & 35.4 & {\bf 22.8} & 27.3 & 32.2\\
With attention & {\bf 36.3}  & {\bf 24.6}  & {\bf 49.9}  & {\bf 25.2}  & 41.8  & {\bf 36.9}  & 22.0  & {\bf 29.7} & {\bf 33.3}\\
\hline
\end{tabular}
\end{center}
\end{table}
\setlength{\tabcolsep}{1.4pt}

Although there are enhancements in multi-task inference, our model can be further improved and tuned for individual tasks. The operation of the AU graph is overfitting to the training set without the support of SAM optimizer. It is less efficient than the FC-Sig module on the validation set when we remove SAM. Regarding the FER task, the prediction of \textit{happy} label, which is a common emotion, is not improved when the attention mechanism is applied.

\section{Conclusion}
In this work, we introduced the attention-based approach to the multi-task learning problem. The idea is to exploit the output of one task to enhance the performance of another task. In particular, our model attends to the AU detection task's result to achieve better output in the FER task. Our result is outstanding compared to the baseline and the cross attention module improves the evaluation metric on the FER task. The experiment demonstrates that facial AUs have a strong relationship with facial expression and this relation can be leveraged to recognize human emotion more efficiently. Additionally, we take advantage of the GCN and the batch normalization to accomplish the AU detection and VA estimation task, respectively, with considerable advancement.

However, in our model, the progress to generate valence and arousal is completely independent of other tasks. In our architecture, except for the image feature, there is no common knowledge between VA estimation and remaining heads. The relation between valence-arousal and other facial attributes is not exploited in this paper. In the future, we plan to study the influence between valence-arousal and other facial behavior tasks.

\section{Acknowledgement}
This work was supported by a National Research Foundation of Korea (NRF) grant funded by the Korean government (MSIT). (NRF-2020R1A4A1019191).

% \clearpage
\bibliographystyle{splncs04}
\bibliography{egbib}
\end{document}